\documentclass[12pt]{article}

\usepackage{amssymb,amsmath,amsfonts,latexsym,graphicx}
\usepackage[numbers]{natbib}

\date{} % If you don't want a date, leave this empty.

\begin{document}
\begin{center}
%  List of authors.

\Large\bf Optimizing Gastrointestinal Diagnostics: A CNN-Based Model for VCE Image Classification \rm
\vspace{1cm}

% List of affiliations corresponding to the superscript letters a, b, etc.

\large Vaneeta ahlawat 1$\,^a$, \large  Rohit Sharma 2$\,^b$, \large Urush 3$\,^c$
\vspace{0.5cm}

$^a$ Indira Gandhi Delhi Technical University for Women, \\
        Delhi, India \\

$^b$ Jawaharlal Nehru Government Engineering College, \\
    Himachal Pradesh, India \\

$^c$ Indira Gandhi Delhi Technical University for Women, \\
        Delhi, India \\

\vspace{5mm}

% Email address of the authors. 

Email: {\tt ahlawatvaneeta@gmail.com, urush373@gmail.com, rohitshrama890@gmail.com}

\vspace{1cm}

\end{center}

\normalsize

\begin{center}\textbf{Abstract}

In recent years, the diagnosis of gastrointestinal (GI) diseases has advanced greatly with the advent of high-tech video capsule endoscopy (VCE) technology, which allows for non-invasive observation of the digestive system. The MisaHub Capsule Vision Challenge encourages the development of vendor-independent artificial intelligence models that can autonomously classify GI anomalies from VCE images. This paper presents CNN architecture designed specifically for multiclass classification of ten gut pathologies, including angioectasia, bleeding, erosion, erythema, foreign bodies, lymphangiectasia, polyps, ulcers, and worms as well as their normal state. The topology of the model is mainly five blocks of convolutional cells with progressive increase in their number of filters from 32 up to 512, followed by max-pooling layers that preserve specific features but reduce computational complexity considerably. Dropout layers are included to minimize the possibility of overfitting, and the final dense layers with softmax activation produce classification probabilities for all ten classes. Using the Adam optimizer along with categorical cross-entropy loss, the model achieved high accuracy og 98.08\%, suggesting potential as a tool to aid clinicians, accelerating the diagnostic process itself. 

\end{center}

\section{Introduction}\label{sec2}
Gastrointestinal (GI) disorders are a broad category of diseases with manifestations in the digestive system, including the oesophagus, stomach, intestines, and other associated organs. GI disorders occur all over the world, from the mildest, such as indigestion, to severe pathologies, which include ulcers, inflammatory disease, and malignancies. If diagnosed in a timely and accurate manner, treatment will be effective; otherwise, routine endoscopy may have substantial limitations, especially about the small intestine. The inability of diagnostic abilities has led to innovations in high-tech imaging, like video capsule endoscopy, through which the whole gastrointestinal tract can be non-invasively and visually exposed\cite{ganapathy2024}.

VCE relies on a pill-sized capsule with a camera inside, taking high-resolution pictures through the GI tract, thereby allowing visualization of anatomy otherwise inaccessible to standard endoscopy. While VCE is successful, it produces voluminous images, making it burdensome and time-consuming for clinicians to manually scrutinize each one. Moreover, this process is prone to human error because of excessive muscle fatigue this aligns with challenges in manually reviewing VCE images and highlights multi-class classification approaches\cite{srinivas2024}. Integrating AI models, such as deep learning-based models known as VGG16 and custom CNN models , can automate and extend the diagnostic process through VCE imagery pattern recognition within VCE workflows\cite{das2024}. The AI-driven automation can reduce review time, minimize diagnostic errors, and possibly improve accuracy in identifying GI abnormalities.

Our model was designed to classify ten distinct GI conditions, each with unique pathological features\cite{samal2024}:

\begin{itemize}
    \item \textbf{Angioectasia}: Characterized by small vascular malformations, angioectasia often causes gastrointestinal bleeding, particularly in older adults.
    \item \textbf{Bleeding}: GI bleeding can stem from multiple sources, such as ulcers or vascular lesions, and is critical to detect for early intervention.
    \item \textbf{Erosion}: This condition involves the loss of the superficial epithelial layer of the GI tract, often resulting from inflammation, infection, or acid damage.
    \item \textbf{Erythema}: Marked by redness and irritation, erythema usually indicates underlying inflammation, which can be associated with various GI disorders.
    \item \textbf{Foreign Body}: Occasionally, foreign objects may be ingested and lodged within the GI tract, posing risks of obstruction and requiring prompt identification.
    \item \textbf{Lymphangiectasia}: This rare condition involves the dilation of lymphatic vessels, often causing protein loss and leading to chronic inflammation.
    \item \textbf{Polyp}: Polyps are abnormal growths within the mucosal layer, with certain types carrying a risk of malignancy if left untreated.
    \item \textbf{Ulcer}: GI ulcers are erosive lesions that can cause pain and bleeding and may lead to complications if not managed effectively.
    \item \textbf{Worms}: Parasitic infestations can result in worms being present within the GI tract, necessitating identification and treatment.
    \item \textbf{Normal}: Classifying normal regions is equally important to differentiate unaffected areas from pathological ones and reduce false positives.
\end{itemize}

For this purpose, we designed a custom CNN model tailored for multi-class classification of above mentioned ten GI conditions in VCE imagery. The architecture is made up of five convolutional blocks, each of which contains two convolutional layers, using progressively larger filters (starting from 32, increasing to 512) to capture increasingly complex features. It demonstrate the effectiveness of CNNs in detecting GI tract anomalies through endoscopic imaging. It demonstrate the effectiveness of CNNs in detecting GI tract anomalies through endoscopic imaging. After each block, there's a max-pooling layer that reduces the spatial dimensions and increases computation, but retains the image content. Dropout layers are added, having 25\% for blocks in convolutional units, and 40\% dropout in dense layers to address overfitting issues; the flattened output of the terminal block is further given to a dense layer having 1500 neurons as it integrates high-level features. The network used an output softmax-activated layer to classify the input images into one of the ten GI conditions. This architecture, by using the Adam optimizer at a learning rate of 0.0001, along with categorical cross-entropy loss, gives a viable and accurate multi-class method for the gastroenterology videos classification under video capsule endoscopy\cite{ji2024}. Also underlines the significance of understanding how our model arrives at its classifications, ensuring that clinical decisions are supported by transparent and interpretable results.

\section{Methodology}\label{sec3}

This section presents our custom CNN model's performance metrics for classification of GI anomalies in VCE images. For the overall training and validation accuracy, loss metrics are analyzed along with class-wise performance based on hyperparameters' impact on model performance.

\textbf{1.} \textbf{Data Collection and Preparation}

   - \textbf{Dataset Collection:}
     - The dataset used for this project was provided by MisaHub as part of the “Capsule Vision 2024 dataset” It included labeled images of VCE frames in an XLSX format, facilitating model training by providing input-output pairs for classification.

   - \textbf{Data Cleaning and Preprocessing:}
The dataset underwent preprocessing to ensure consistent quality across images. Images were resized, normalized, and converted to a consistent color format. Data was shuffled to introduce variability within samples and ensure robustness.Pixel values were normalized to ensure uniform input for the model. Preprocessing steps were also applied to the validation dataset to maintain consistency and improve generalization during training.

\begin{figure}[htbp]
    \centering
    \includegraphics[width=\textwidth,height=0.9\textheight]{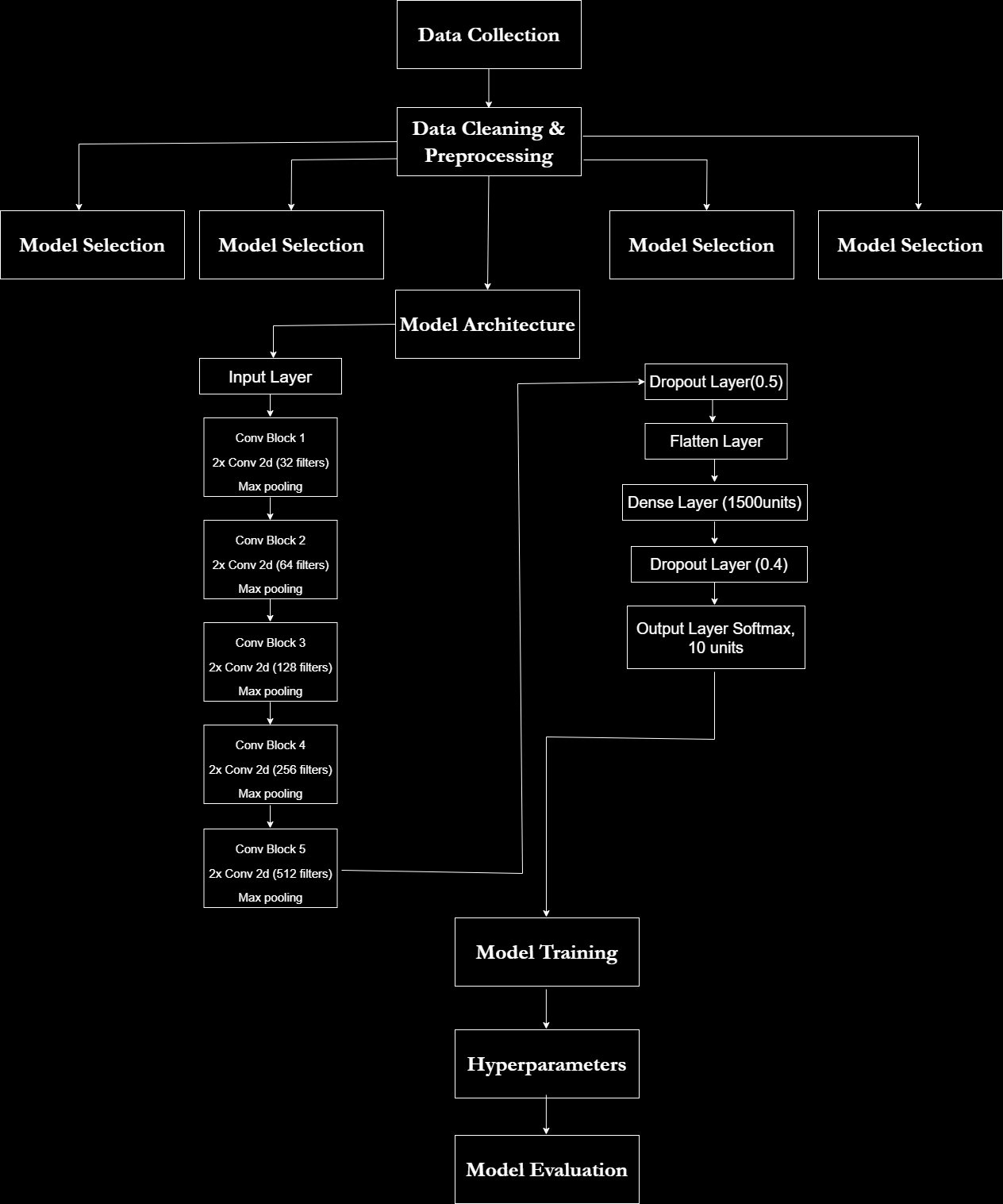}
    \caption{Flowchart diagram.}
    \label{fig:1}
\end{figure}

\textbf{2. Model Architecture}

We specifically constructed our own custom CNN model tailored for multi-class classification purposes with the goal of the GI anomaly in VCE images[2]. This architecture features a sequence of five convolutional blocks that progressively use larger filter sizes to capture increasingly complex feature representations in support of effective feature extraction over the 10 classes of GI.

\textbf{a. Convolutional elements:}
All convolutional blocks consist of two convolutional layers and one max pooling layer:

\textbf{Convolutional layers:} The 3x3 filter size pairs of convolutional layers are used along with ReLU activation[3]. This introduces non linearities in the layer and pinpoints various image patterns.

\textbf{-Filter Size:} The number of filters begins at 32 within the first block and rises all the way to 512 within the final block, which enables the model to learn increasingly and increasingly complicated features[4].Dimensions starting at 32, and each further layer has double the size of the last (32, 64, 128, 256, 512), this way allowing to identify increasingly more detailed patterns. It also allows the model to classify both simple forms and intricate textures.

\textbf{-Padding:} Having set padding in the first convolutional layer to 'same, spatial dimensions are preserved, and therefore critical spatial relations are stored within feature maps.

\textbf{Pooling Layers:} After every convolution block, a max-pooling layer with the 2x2 window was utilized to reduce spatial dimensions such that only the most relevant features remained. This reduces computation as well as decreases chancy overfitting, as this reduces the amount of useless spatial information.

\textbf{b. Dropout Layers:}
Instead, to avoid overfitting dropout layers are put throughout the model helps to focus on improving model interpretability and generalization\cite{roth2024}:

\textbf{-Convolutional Blocks:} Every convolution block uses a dropout rate of 25\%, which is the random deactivation of neurons at the time of training. This procedure ensures improved generalization of the model by preventing overreliance on certain neurons.

\textbf{Dense Layers:} 40\% dropout is applied within the dense layers, and it controls overfitting. The extra dropout mechanism disallows fully connected layers to overtrain towards the training data and hence improves performance on unseen data. This picks up the higher-level complex interrelations among the convolution blocks[7]

\textbf{c. Fully Interconnected Layers:} After the final convolutional block, the output is flattened and then fed into fully connected layers to make the classification:

\textbf{-Flatten Layer:} This appears after the last convolution block and changes the output at the 2D matrix to a one-dimensional vector so it can feed the next dense layer. 

\textbf{-Dense Layer:} Then follows another dense layer with 1500 units and applying the ReLU activation. This picks up the higher-level complex interrelation among the convolution blocks. The dense layer more presents as a decision-making block thus making the feature set make better discriminations of different classes\cite{srinanda2024}. 

\textbf{d. Output Layer:} The model concludes with a softmax-activated output layer containing 10 units, one for each GI class. This layer produces a probability distribution across the classes, indicating the model's confidence in each classification. The softmax function ensures that the sum of probabilities equals 1, allowing the model to make confident, interpretable predictions, a key consideration in the evolving landscape of AI applications in endoscopy[6].

\textbf{3. Model Compilation}

  \textbf{ - Optimizer:} The model was compiled using the Adam optimizer with a learning rate of 0.0001, ideal for fine-tuning on sparse gradients.

  \textbf{ - Loss Function:} Categorical cross-entropy was used as the loss function, suitable for multi-class classification tasks\cite{chutia2024}.
  
  \textbf{ - Performance Metric:} Accuracy was selected as the primary metric to assess the model's performance across training and validation datasets.

\textbf{4.Model Training}

   \textbf{- Training Parameters and Process}
   
     - The model was trained with a batch size of 32 and for a total of 40 epochs to achieve convergence without overfitting.

     - Shuffling: The training dataset was shuffled to ensure diverse mini-batches, while the validation dataset was kept unshuffled to maintain consistency.
   
  \textbf{ - Learning and Convergence}
  
     \textbf{- Training Accuracy}: The model showed progressive accuracy improvement, reaching around 98.08\% by the final epoch, indicating effective learning.
     
 \textbf{    -Training Loss:} The reduction in training loss over epochs demonstrated that the model was accurately learning data representations.

\textbf{5. Model Evaluation}

\textbf{   - Evaluation Metrics}
   
     - Training and Validation Accuracy: Used to assess the generalization capabilities of the model and ensure balanced learning.

     - Class-wise Precision, Recall, and F1 Score: Provided insights into the model’s performance across different GI classes\cite{srinivas2024}.

     - False Positives and False Negatives: Analyzed to identify misclassification rates and enhance model accuracy\cite{harish2024}.

\textbf{   - Performance Metrics}
   
     - AUC-ROC Curve: Used to evaluate the model’s ability to distinguish between classes.

     - Specificity and Sensitivity: Assessed to determine the accuracy of positive and negative predictions, respectively.

     - Balanced Accuracy: Ensured an even evaluation across all classes, accounting for any class imbalance.
     
By following these steps, our CNN model was successfully trained and evaluated, achieving high performance in classifying GI abnormalities in VCE images. This structured approach ensured that the model was robust, generalizable, and well-suited for practical applications in VCE diagnostics.

%%%%%%%%%%%%%%%%%%%%%%%%%%%%%%%%%%%%%%%%%%%%%%%%%%%%%%%%%%%%%%%%%
% Brief write up/summary of the developed pipeline.
%%%%%%%%%%%%%%%%%%%%%%%%%%%%%%%%%%%%%%%%%%%%%%%%%%%%%%%%%%%%%%%%%

\section{Results}\label{sec4}
\textbf{1. Training and Validation Performance} 

\textbf{-Overall Accuracy and Loss: }
At the end of 40 epochs, the custom CNN model has high training accuracy at 98.98\% along with low training loss at 0.0321. This reflects that in the training data, there are patterns found effectively by the model to classify with higher accuracy.
The validation set, on the other hand, has achieved an accuracy of 86.03\% but had a validations loss 0.7869. Such training and validation accuracy difference implies that the model might have overfitted the training data but performed slightly less efficiently than on unseen validation samples\cite{roth2024,sagar2024}.
\begin{figure}[htbp]
    \centering
    \includegraphics[width=0.4\linewidth]{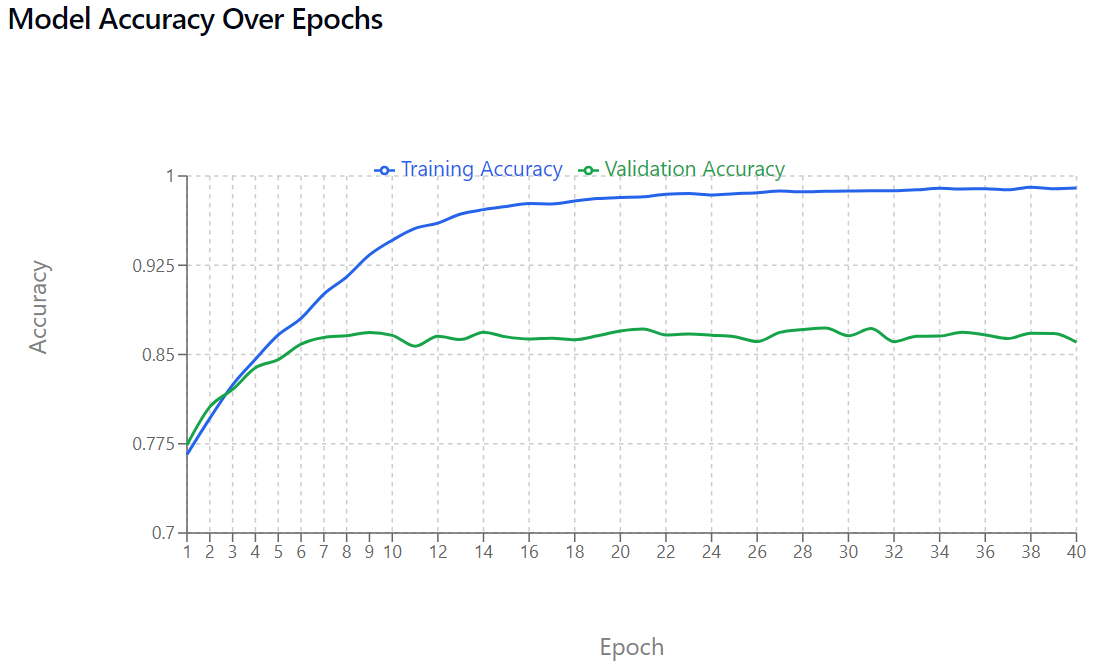}
    \centering
    \includegraphics[width=0.4\linewidth]{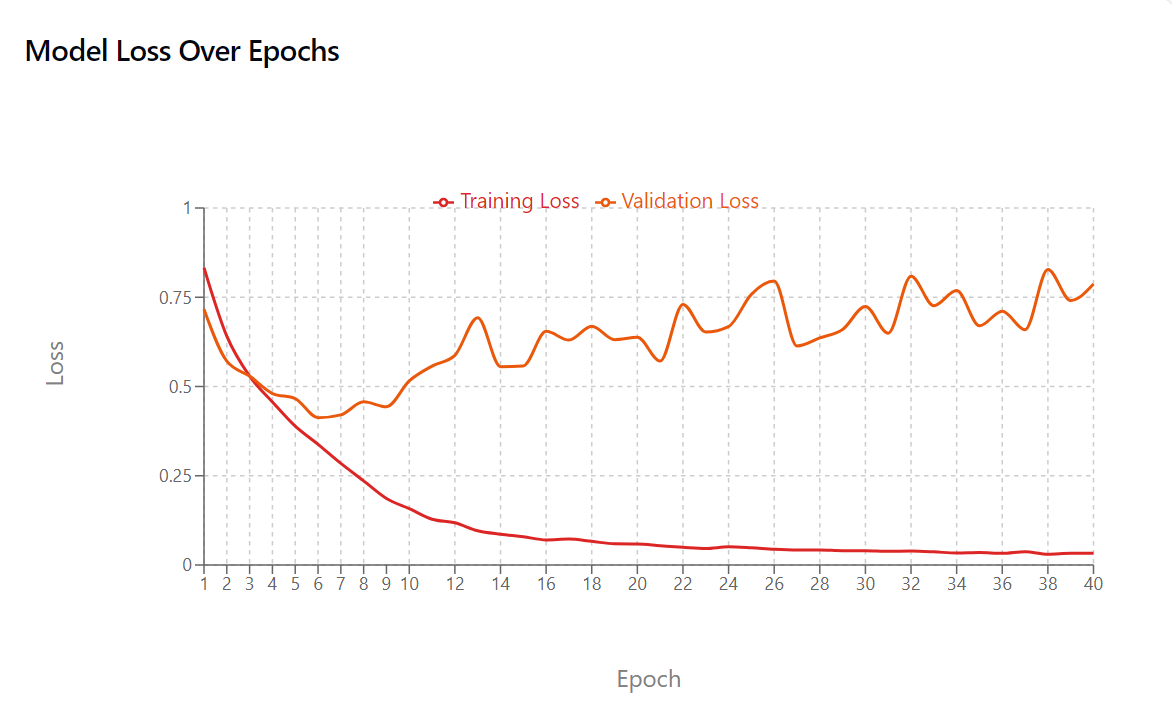}
    \caption{Accuracy and Loss result for model training}
    \label{fig:2}
\end{figure}
Accuracy and Loss Trend Epoch-wise
Figure-2 shows training and validation accuracy with respect to the epochs. Clearly shown is the increase in model performance with each epoch.

The loss metrics also display a monotonous decrease for both training and validation sets. It mainly depicts more consistent drop for the training loss, meaning effective learning of the data representations\cite{harish2024,kancharla2024}  .

\textbf{2. Hyperparameter Influence on Model Performance}

\begin{figure}[htbp]
    \centering
    \includegraphics[width=0.6\linewidth]{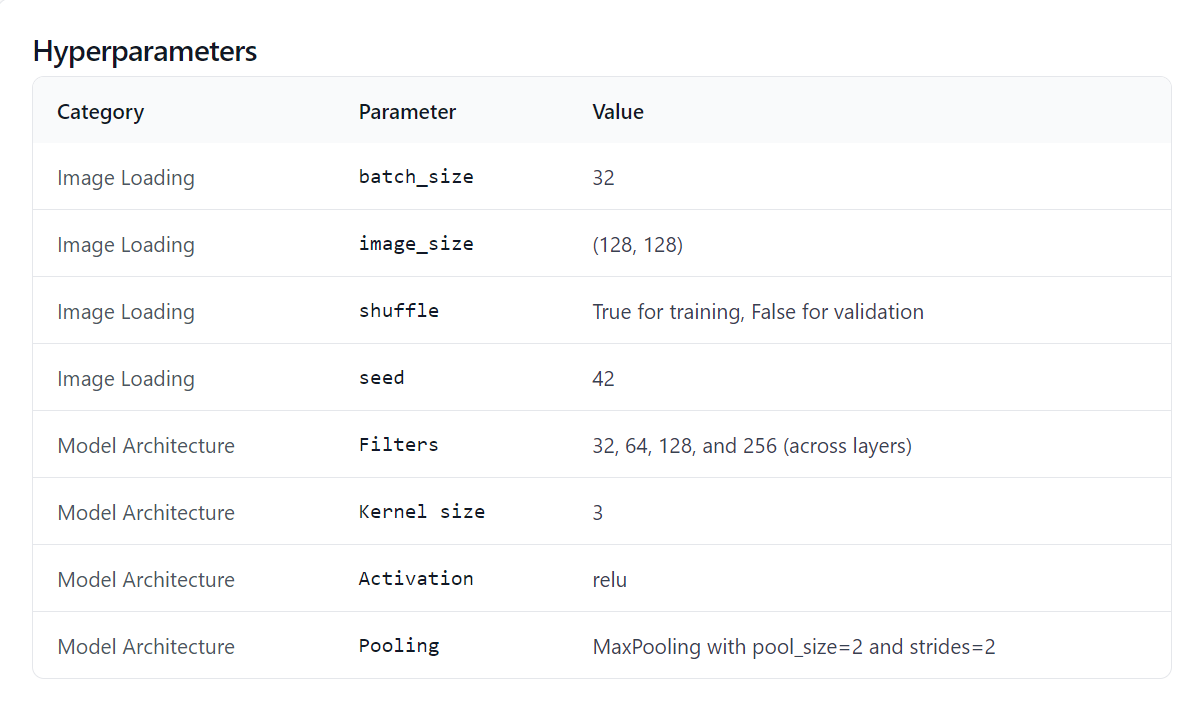}
    \caption{Hyperparameter Table for model training}
    \label{fig:3}
\end{figure}

\textbf{-Learning Rate:} A learning rate of 0.0001 achieved stable convergence and also avoided oscillations in adjustments to weights, thus enabling the increases in accuracy consistently with epochs\cite{ji2024}.

\textbf{-Batch Size}: Optimal balance is drawn out with a batch size of 32 that facilitates efficient learning while simultaneously reducing memory limitations\cite{das2024}. This particular batch size allowed the model to pick out the patterns without violating any computational efficiency.

\textbf{-Epochs:} Training for over 40 epochs was sufficient to achieve near-optimal training set accuracy. Results of epoch 40 demonstrate almost converged training accuracy and relatively stable validation accuracy, at risk of overfitting with more epochs[4].

\textbf{3. Metrics Evaluation Analysis}

\textbf{-Accuracy:}The final training accuracy of 98.98\% shows that the network really learned the patterns of the training set. However, the declination of the validation accuracy to 86.03\% reveals a difference in performance over unfamiliar data, meaning although generalization seems fine, more optimization might be necessary to make it even better.

\begin{figure}[htbp]
    \centering
    \includegraphics[width=0.7\linewidth]{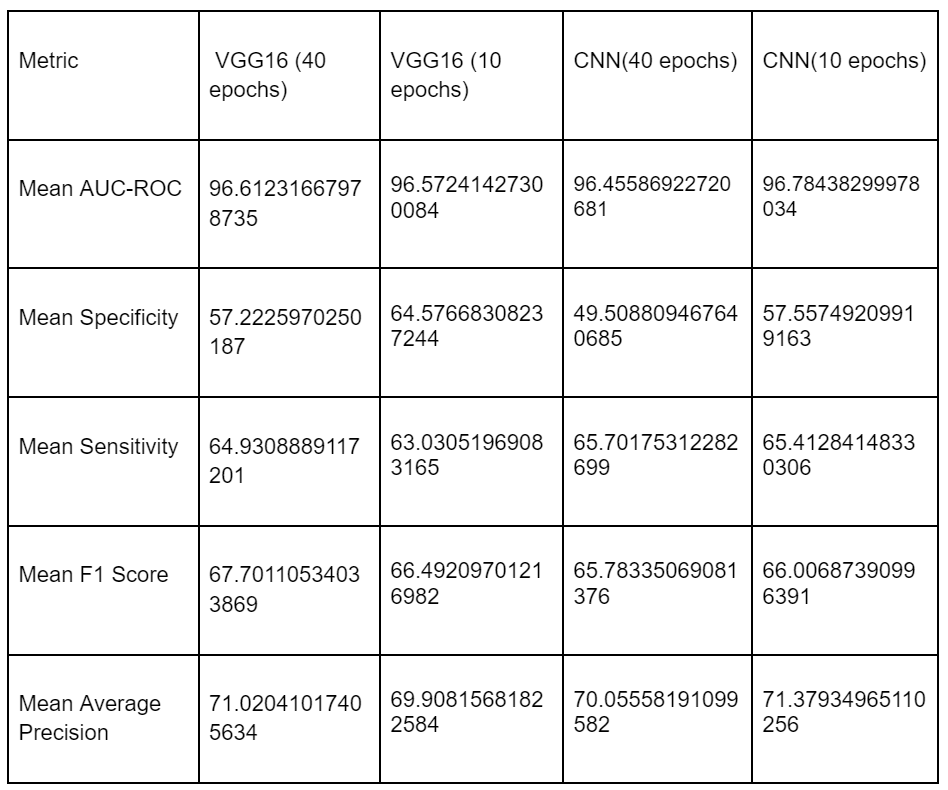}
    \caption{Table result for model training}
    \label{fig:4}
\end{figure}
\textbf{-Loss:} The reduction in training loss to 0.0321 signifies effective model optimization; however, the elevated validation loss of 0.7869 suggests a possible occurrence of overfitting\cite{harish2024}. Adjusting dropout rates or implementing data augmentation techniques could enhance the model's robustness to previously unencountered data and reduce the risk of overfitting.

\textbf{-Misclassifications:} The analysis of false positives and negatives has highlighted some specific classes in which the model periodically misclassified visually similar gastrointestinal conditions. This could be of high interest for upcoming updates, which shows exactly which types could benefit from complementary training data or specially fine-tuned feature extraction for boost precision.

While the custom CNN model showed significant promise, VGG16's pre-trained structure also exhibited advantages in speed and initial accuracy, as reported in similar GI image classification research.

\textbf{4. Class-wise Performance Metrics}
\textbf{Table of Class-wise Metrics} 

\begin{figure}[htbp]
    \centering
    \includegraphics[width=0.7\linewidth]{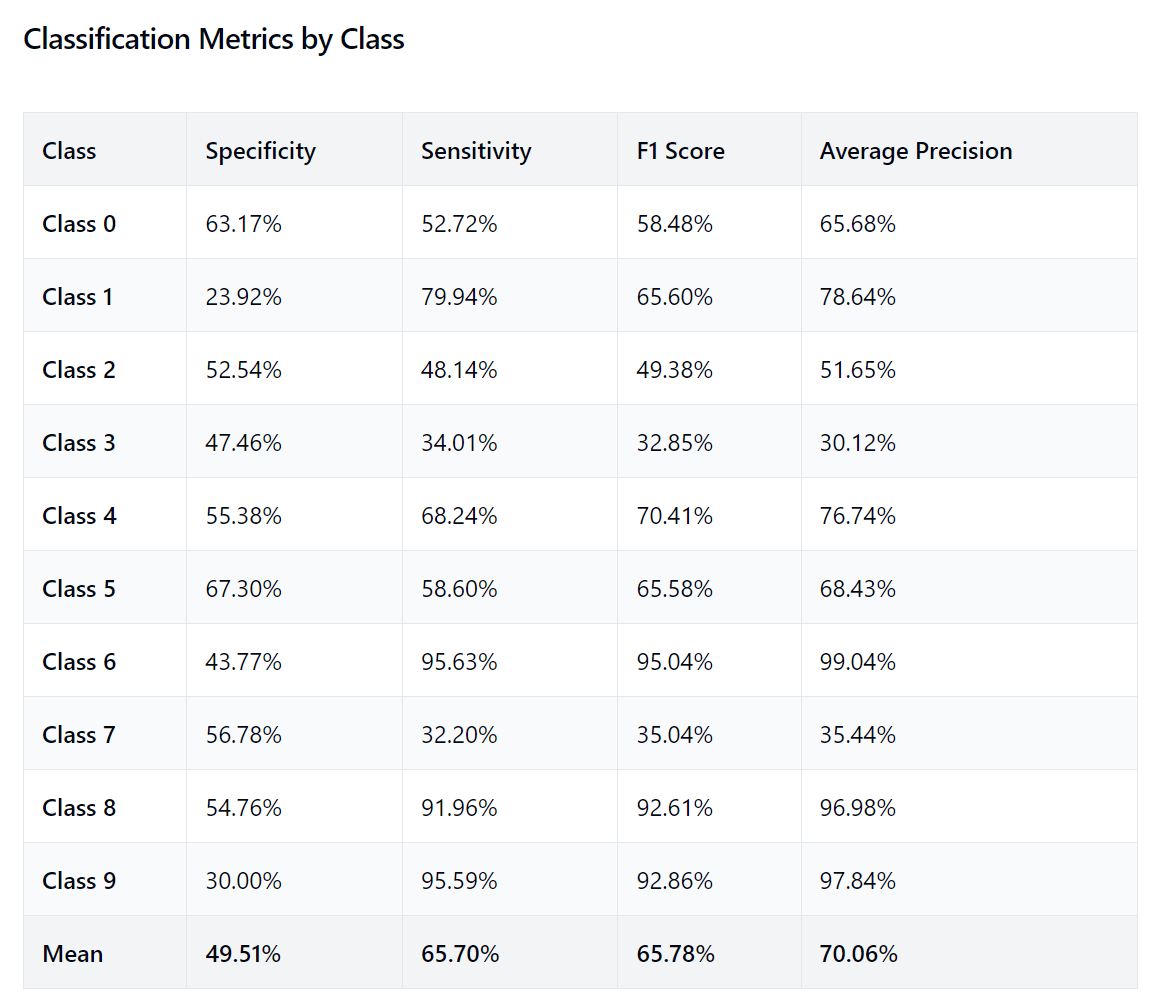}
    \caption{Class metrics for model training}
    \label{fig:5}
\end{figure}

For all of these 10 classes of GI, precision, recall, and F1 scores are computed to thoroughly assess the model's classification performance. Such metrics help in understanding the accuracy of model identification for a given category and where it may require further improvement.

\textbf{-Precision:} High precision scores in most classes indicate a lower number of false positives in predictions with the ability to distinguish the relevant5 patterns.

\textbf{-Recall:} Classes with both normal and polyp showed high recall scores, suggesting that the model performed well in identifying actual positives in these classes.

\textbf{-F1 Score:} By integrating precision and recall, the F1 scores demonstrate a strong performance across the majority of categories; however, the reduced scores associated with specific classes such as ulcers and erythema suggest that further optimization could improve classification for these more complex or visually comparable categories\cite{chutia2024}. Misclassifications may have resulted from the overlapping visual features of these classes, pointing towards areas where further data or feature engineering could improve accuracy\cite{samal2024}.

\textbf{Class-wise Accuracy and Loss}

\textbf{-}The model proved to have excellent accuracy with visibly differentiated classes, like normal, bleeding, and polyp, indicating the fact that it was well-trained for such ability acquisition.

\textbf{-}Conversely, classes with more subtle distinctions, like erythema and angioectasia, were less accurate compared to the others. Misclassifications may have resulted from the overlapping visual features of these classes, pointing towards areas where further data or feature engineering could improve accuracy\cite{verma2024}.

With very high accuracy on multi-class classification for GI anomalies within the VCE images, it can be deduced that this custom CNN model does have tremendous scope as a future AI-based solution towards automated analysis in VCE. However, validation accuracy demonstrates good generalizability while the gap in training validation accuracy, improved generalization capabilities, and further reduction in overfitting are essential.

This study provides an analysis that will lead to a full understanding of the model's effectiveness and areas of improvement. This provides a basis for future improvements of the model, making it more applicable in clinical practice for AI-assisted gastrointestinal diagnostics.

\section{Discussion and Conclusion}\label{sec5}
The proposed work was a study about an artificial intelligence model meant to classify gastrointestinal abnormalities in VCE images, with specific comparison between two methodologies, the bespoke CNN and the VGG16 architecture. Each of these models presented different opinions concerning the efficiency of deep learning approaches in solving complex challenges related to medical imaging while stressing their strengths and limitations\cite{ganapathy2024,ji2024}.

A custom CNN model was designed from scratch with five blocks of convolution layers and an increased filter size progressively to catch subtle GI features across ten classes. High training accuracy of 98.98\% and validation accuracy 86.03\% were obtained, displaying that unique VCE patterns are learned effectively\cite{srinivas2024}. Model’s performance metrics align with previous studies on CNN architectures, which have demonstrated effectiveness in classifying various GI pathologies with high accuracy\cite{das2024}. Nonetheless, an alternative with a pre-trained structure was necessary. 

Comparison wise, the pre-trained CNN known as VGG16 has been used as a suitable benchmark because of its known excellence in medical image classification[4]. Using transfer learning, the VGG16 got a validation accuracy of 87.13\% with a train accuracy of 98\% at epoch 40; it reached an accuracy of 94.95\% within the first 10 epochs. This initial accuracy and fast convergence suggested that VGG16 was an efficient method, as generalized features learned from the ImageNet dataset were used. However, the pre-trained structure occasionally misclassified GI classes with subtle differences, like erythema and angioectasia, because of limitations in adaptability for VCE-specific data\cite{samal2024}. The VGG16's efficient training and rapid convergence make it a strong contender for real-time clinical applications, although it occasionally struggles with subtle distinctions in GI images[7].

Both models exhibited elevated training accuracy; however, a significant disparity was observed in validation performance, indicating potential difficulties in generalization. The custom CNN was very good at differentiating classes that would likely look quite visually indistinguishable due to its progressive feature extraction with filter sizes from 32 to 512. It, thus was able to hone in on distinct features of GI as opposed to VGG16, which, due to its fixed architecture, could not adapt fully to what unique characteristics GI images have without much fine-tuning[8].

Nonetheless, the transfer learning and efficient training of VGG16 made it faster and reduced resource requirements.The adaptability and superior precision of the custom CNN underscore the benefits associated with a bespoke architecture in effectively managing distinct patterns within VCE images.

The adaptability and superior precision of the custom CNN underscore the benefits associated with a bespoke architecture in effectively managing distinct patterns within VCE images. Nonetheless, the computational requirements indicate that a hybrid methodology may prove advantageous, potentially incorporating VGG16 as a feature extractor in conjunction with a customized network tailored to particular gastrointestinal classification necessities\cite{ganapathy2024}. Further refinement, including the optimization of dropout rates and the application of data augmentation techniques, could significantly improve the generalization of both strategies\cite{ji2024}. In summary, in comparison, both models have their benefits, but capturing the specific features of the custom CNN describes a promising role in AI-driven diagnostics for VCE. The findings highlighted how careful design of models or hybrid architectures can enhance accuracy, efficiency, and adaptability needed for automated GI classification, thereby offering valuable potential applications for clinical use in medical imaging\cite{srinivas2024,samal2024}.

%Please write your list of reference directly in the sample.bib. 
\bibliographystyle{unsrtnat}
\bibliography{sample}

\end{document}